\documentclass[submission,copyright,creativecommons]{eptcs}
\pdfoutput=1
\providecommand{\event}{Doctoral Consortium - ICLP 2022} % Name of the event you are submitting to
\usepackage{breakurl}             % Not needed if you use pdflatex only.
\usepackage{underscore}           % Only needed if you use pdflatex.
\usepackage{subcaption}
\usepackage{graphicx}
\usepackage{graphics} % for pdf, bitmapped graphics files
\usepackage{epsfig} % for postscript graphics files
\usepackage{todonotes}

\usepackage{listings}
\usepackage{asplisting}
\lstnewenvironment{asp}[1][]
{
	\lstset{
		language=asp,
		showstringspaces=false,
		formfeed=\newpage,
		tabsize=4,
		commentstyle=\color{asparagus},
		basicstyle=\ttfamily\scriptsize,
		keywordstyle=\color{blue},
		numbers=left,
		breaklines=true,
		literate={~} {$\sim$}{1},
		#1
	}
}
{
}

\newcommand{\ttt}[0]{\texttt}

\title{An ASP Framework for Efficient Urban Traffic Optimization}
\author{Matteo Cardellini
\institute{DIBRIS, 
Università degli Studi di Genova\\
Genova, Italy}
\institute{Politecnico di Torino\thanks{The author is a PhD Student at the Italian National PhD Programme in Artificial Intelligence \url{https://www.phd-ai.it/}}\\
Torino, Italy}
\email{matteo.cardellini@polito.it}
}
\def\titlerunning{An ASP Framework for Efficient Urban Traffic Optimization}
\def\authorrunning{M. Cardellini}
\begin{document}
\maketitle

\begin{abstract}
Avoiding congestion and controlling traffic in urban scenarios is becoming nowadays of paramount importance due to the rapid growth of our cities' population and vehicles. The effective control of urban traffic as a means to mitigate congestion can be beneficial in an economic, environmental and health way. In this paper, a framework which allows to efficiently simulate and optimize traffic flow in a large roads' network with hundreds of vehicles is presented. The framework leverages on an Answer Set Programming (ASP) encoding to formally describe the movements of vehicles inside a network. Taking advantage of the ability to specify optimization constraints in ASP and the off-the-shelf solver \textsc{Clingo}, it is then possible to optimize the routes of vehicles inside the network to reduce a range of relevant metrics (e.g., travel times or emissions). Finally, an analysis on real-world traffic data is performed, utilizing the state-of-the-art Urban Mobility Simulator (\textsc{SUMO}) to keep track of the state of the network, test the correctness of the solution and to prove the efficiency and capabilities of the presented solution.
\end{abstract}

\section{Introduction}
\paragraph{Problem Description.} At the end of the 21st century, the world population is expected to increase to 10.9 Billion, adding more than 3 Billion people to the current population \cite{roser2013world}. This huge growth, which will directly translate in a higher number of vehicles roaming the streets of our cities, demands improvements in the transport infrastructure and a better utilization of our roads for the purpose of avoiding congesting the network. Traffic jams have a negative impact on safety and fuel consumption, which directly translates to a higher cost for drivers and health issues for residents near highly trafficked roads, caused by bad air quality and noise pollution \cite{van2004driving}. 

\paragraph{Overview of existing literature.} One of the most common methods in the literature to optimize traffic flow in road networks is to schedule traffic light switching phases, or \textit{signal phase plans (SPPs)}, with the aim of minimizing delay and avoiding wasting time at intersections \cite{papageorgiou2003review, dotoli2006signal}. The well-known real time adaptive traffic control system SCOOT \cite{bretherton1990scoot}, for example, makes use of this methodology and is now used extensively throughout the United Kingdom. Unfortunately, managing only the switching phases of traffic lights has some limitations: the only metric which is controllable and optimizable is the waiting time at intersections (which has a direct impact on the total travel time of vehicles) but is not straightforwardly expandable to consider other metrics (i.e., pollution, risk, fuel consumption, etc). Moreover, the (\textit{macroscopic}) point of view of traffic lights, which manages the flow of traffic modelling incoming and outgoing lanes as queues of vehicles, does not allow for a more detailed (\textit{microscopic}) consideration of the single vehicles and their routes inside the network. Microscopic simulation models have been largely discarded in the literature due to their high complexity and low scalability. In this paper, we leverage state-of-the-art Artificial Intelligence techniques, coupled with domain-dependent optimizations, to model the traffic flow of hundreds of vehicles inside large European cities from a \textit{microscopic} point of view.

The use of Artificial Intelligence techniques in road transportation has already been found to be efficient in optimizing traffic flow \cite{abduljabbar2019applications,miles2006potential}.  For instance, \cite{vallati2016efficient} introduced a \textit{mixed discrete-continuous planning} \cite{fox2006modelling} approach for reducing congestion through a \textit{macroscopic} point of view. In \cite{chrpa2016automated} instead, a \textit{temporal planning} approach was used for managing traffic, now through a \textit{microscopic} prospective, with the intention of reducing air pollution and respect air quality limitations. Other instances of urban traffic problems solved with automated planning technologies can be found in \cite{cenamor2014planning}. Even if automated planning has been beneficial in efficiently solving several real-world problems in transportation \cite{cardellini2021station, ramirez2018integrated}, the main point of failure is the ability to scale in the presence of large number of vehicles. 

\paragraph{Goal of the research.} Arguably, the purpose of optimizing the flow of traffic inside a road network lies not in finding the best possible route for every vehicle, but instead in finding the best combination (schedule) of routes for all the vehicles in the network. For this reason, the approach presented in this paper relies mainly on an Answer Set Programming (ASP) \cite{DBLP:journals/ngc/GelfondL91,DBLP:journals/amai/Niemela99,baral2003} encoding of the problem. ASP has already proved to be an effective tool for solving several practical scheduling problems \cite{DBLP:conf/lpnmr/DodaroM17,DBLP:journals/tplp/DodaroGGMMP21,DBLP:conf/ruleml/CardelliniNDGGM21}. In particular, in \cite{DBLP:conf/ecai/EiterFSS20}, an ASP based solution was presented for modelling an abstract \textit{mesoscopic} flow model (a middle-ground approach between \textit{macroscopic} and \textit{microscopic}) and a strategy for generating traffic lights SPPs. In this paper, the ASP encoding will have the goal to find the best route (according to some metric) for every new vehicle that enters the network using a relaxed version of the network and traffic rules. After computing the best route in the relaxed system, the real metrics and performance are computed using the microscopic traffic simulation tool SUMO \cite{DBLP:conf/itsc/LopezBBEFHLRWW18} which will instead consider all the difficulties of the real system. Our proposed framework will behave as a \textit{Centralized Urban Traffic Controller} (CUTC): we will suppose that the centralized controller knows the position of every vehicle inside the network and is tasked to find the best possible route (according to some metric) for new vehicles which enter the network.

\paragraph{Structure of the paper.} Section \ref{sec:current-status} discusses the architecture of the proposed framework, the domain specific optimizations which allow the system to scale to manage a high number of vehicles, and the ASP encoding which is used under the hood to schedule the routes inside the network. Section \ref{sec:results} shows how the proposed framework compares with other  approaches and how the system scales with respect to the number of vehicles which are inside the network. Section \ref{sec:future} closes the paper by discussing open issues of the framework and how these can be tackled in future work.

\section{Current status of the research} \label{sec:current-status}
\begin{figure}
    \centering
    \includegraphics[width=\textwidth]{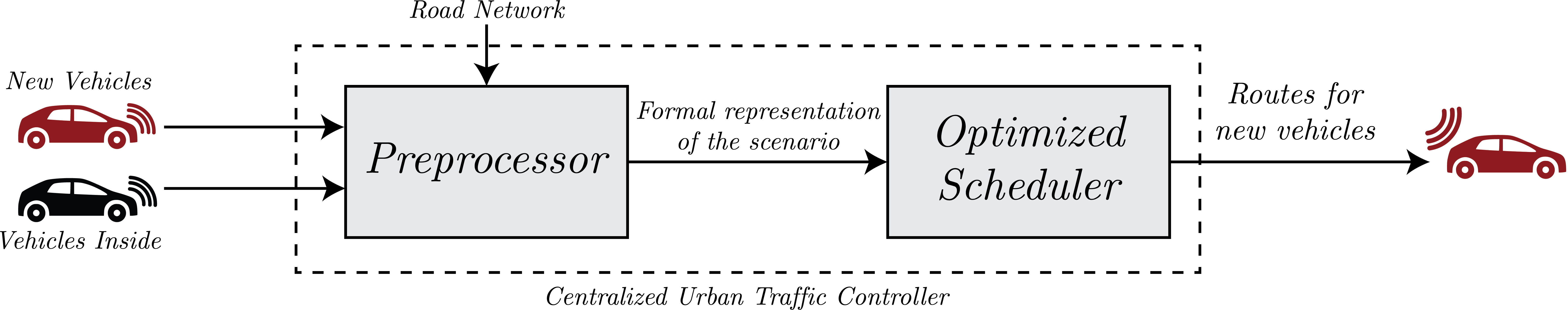}
    \caption{Architecture of the proposed framework}
    \label{fig:architecture}
\end{figure}

\subsection{Architecture} In the last decade, driven by the concept of \textit{smart cities}, several communications systems between vehicles and traffic components (i.e., traffic lights, traffic controllers, etc) have been developed \cite{emmelmann2010vehicular}. As previously stated, the proposed framework purpose is to act as a CUTC, which has knowledge of the position of vehicles inside the network and is tasked to find an optimal route (according to some metric) for new vehicles which approach the network.  In the proposed framework, the system is able to communicate with vehicles through \textit{Vehicular Ad Hoc Networks} (VANETs) and fetch data about vehicles (i.e., their position and route inside the network). Figure \ref{fig:architecture} shows the architecture for the proposed framework, which is composed of the following components: 
\begin{itemize}
    \item a \textit{Preprocessor} is dedicated to build an internal model of the road network, abstracting part of it (i.e., joining small streets together and easing intersections in roundabouts) and to compute preliminary results (i.e., list of possible routes for every vehicle, street's enter and exit times ranges for every new vehicle) in order to guide the computation and avoiding having the \textit{Optimized Scheduler} searching solutions which are already known to be unpromising. This component takes as input the network structure, the list of incoming new vehicles and the position of all the vehicles which have already a route inside the network (through VANETs). The preprocessor outputs then a simplified formal representation describing the network, the possible routes of new vehicles and the status of the vehicles inside the network.
    
    \item an \textit{Optimized Scheduler} receives a formal representation of the scenario, which has been constructed by the preprocessor, and finds an optimal route for each vehicle which is approaching the network. These optimized routes are then communicated to the approaching vehicles through the VANETs. 
    
\end{itemize}
More in depth, in the proposed approach the formal representation of the traffic scenario is specified using an Answer Set Programming encoding and the \textit{Optimized Scheduler} is a wrapper of the general purpose solver \textsc{Clingo} \cite{DBLP:conf/iclp/GebserKKOSW16}.

\subsection{Domain Specific Optimizations} \label{sec:dso}

Before describing the ASP Formulation of the encoding, it is important to elaborate on how the \textit{Preprocessor} reasons upon the network and the vehicles that occupy it as a means of simplifying the solving process, excluding solutions which are known beforehand to be infeasible, or not optimal:
\begin{itemize}
    \item In the proposed architecture, the origin and destination of each vehicle which approaches the network is known a priori. For this reason, it is possible to compute beforehand, for all the entering vehicles, all the possible (acyclic) paths in the graph network which connects their source and target streets. In large and complicate maps, this would result in hundreds of possible routes. For this reason, the \textit{preprocessor} groups similar routes together and then takes a couple from the shortest of each group. This will allow the scheduler to have the most different (shortest) routes in order to deal with traffic.
    \item Since the scheduling has to be time-dependent, it is important to discretize the time in steps which are sufficiently small to capture real traffic nuances but are large enough to still be able to schedule in a reasonable time. In our experiment, the discretization step was chosen to be of $5s$. 
    \item A single roundabout is composed of several very small streets which connect all the intersections of incoming and outgoing streets. The small streets result in a high penalty for vehicles that need to cross the roundabout, since every small street of the roundabout must be run in the discretization step ($5s$). For this reason, the preprocessor joins together all the streets which connects all incoming and outgoing streets of the roundabout, creating longer streets which do not penalize crossing the roundabout so heavily. Since streets are now grouped together, a single street is accounted multiple times in the network, and it is up to the scheduler to respect the total capacity of the roundabout.
    \item Knowing all the routes of vehicles inside, and their expected position at every point in time (obtained from the scheduler when the vehicles were approaching the network), it is possible, for every route that a new vehicle could run, to compute beforehand a range on the timings of entrance and exit in every street of the route. Intuitively, the minimum time in which a vehicle enters and exits a street in a route can be computed as if all traffic were removed, meaning the vehicle is not slowed down and the streets are run at maximum speed. A maximum exit time, instead, can be computed by analysing how many vehicles would be present in the street at its minimum entry time: (i) if the vehicles inside are less than the street capacity then we can consider the speed inside the street to be inversely proportional to the number of vehicles in the street, (ii) if the vehicles congest the street than we can compute a maximum exit time by considering how much it will take for the queue to move to the controlled vehicle.
\end{itemize} 

\subsection{ASP Formulation}
The main core of the proposed architecture lies in the ASP formulation of the traffic scenario. Here, the ASP encoding is presented, based on the input language of \textsc{Clingo}. In the following, we assume the reader is familiar with syntax and semantics of ASP; for details about syntax and semantics of ASP programs, the reader is referred to \cite{DBLP:journals/tplp/CalimeriFGIKKLM20}. Firstly, the network-specific optimized data model produced by the preprocessor is introduced, afterwords the static encoding which models the behaviour of traffic flows inside the map will be discussed.

\paragraph{Data Model.} The data model is composed of the following atoms:
\begin{itemize}
    \item \ttt{streetOnRoute(S,R,MIN,MAX)} which models the fact that street \ttt{S} composes route \ttt{R}. \ttt{MIN} and \ttt{MAX} represent the minimum and maximum time a vehicle, which starts  to run through route \ttt{R} at $t=0$, is expected to enter street \ttt{S}.
    
    \item \ttt{link(S1,S2)} specifies that it is possible for a vehicle to move from street \ttt{S1} to \ttt{S2},
    
    \item \ttt{vehicle(V,T)} defines the presence in the map of a vehicle \ttt{V} of type \ttt{T}, which can be \ttt{1} if the vehicle is \emph{controlled}, meaning that it is a new vehicle for which a route has yet to be found, or \ttt{0} if the vehicle is \emph{simulated}, meaning that a route is already been set and the system need only to keep track of their presence in the map.
    
    \item \ttt{origin(V,FROM)} designates \ttt{FROM} as the street the vehicle \ttt{V} is at the time of the planning. If a vehicle is \emph{controlled}, the origin coincides with the first street the vehicle will step into when entering the map. Similarly, \ttt{destination(V,TO)} specifies that \ttt{TO} is the final street which will bring the vehicle \ttt{V} outside the map.
    
    \item \ttt{possibleRouteOfVehicle(V,R)} signals that a vehicle \ttt{V} in order to move from its origin to its destination can follow a route \ttt{R}. If the vehicle is \emph{simulated} then it will only have one possible route available, which is the one found when the vehicle entered the network, conversely if a vehicle is \emph{controlled} this atom will specify a subset of all the possible routes from origin to destination, as discussed in the previous section.
    
    \item \ttt{time(T)} specifies the time unit of the scheduling, ranging from $0$ to the maximum horizon in which all vehicles have left the street. As previously stated, in the proposed approach the time step has been chosen to be of $5s$. 
    
    \item \ttt{capacity(S,N)},  \ttt{heavyTrafficTravelTime(S,T)}, \ttt{mediumTrafficTravelTime(S,T)}, and \ttt{lightTrafficTravelTime(S,T)} indicate respectively the capacity of street \ttt{S} and the times needed to run the street in cases of heavy traffic (15 km/h), medium traffic (30 km/h), low traffic (45 km/h) based on the length of the street. \ttt{maxTrafficTravelTime(S,T)}, similarly, models the time it would take to free the street in cases of congestion in which the street is at it maximum capacity.
    
    \item \ttt{heavyTrafficThreshold(S, MIN, MAX)}, \ttt{mediumTrafficThreshold(S, MIN, MAX)} and the atom \ttt{lightTrafficThreshold(S, MIN, MAX)} specify the range $(\ttt{MIN}, \ttt{MAX}]$ of vehicles in street \ttt{S} which characterize respectively heavy, medium and low traffic of the previous point.
    
    \item \ttt{enter(V,S,IN)} and \ttt{exit(V,S,OUT)} specify that the \emph{simulated} vehicle \ttt{V} will enter and exit street \ttt{S} at time \ttt{IN} and \ttt{OUT}, respectively. These atoms are constructed from the scheduling solution found previously.
    
    \item \ttt{roundabout(R, C)} specifies the existence of a roundabout \ttt{R} with maximum capacity (in all its streets) of \ttt{C}. \ttt{streetInRoundabout(SS,R)} indicates that the (simplified) street \ttt{SS} belongs to the roundabout \ttt{R}.
    
    \item \ttt{cost(R,V,N)} specifies that running on route \ttt{R} for vehicle \ttt{V} has a cost of \ttt{N}.
    
\end{itemize}

\begin{figure}[t!]
    \centering
    \begin{asp}
$\label{enc:guessRoute1}$ 1 {solutionRoute(V, R): possibleRouteOfVehicle(V, R)} 1 :- vehicle(V, 1).
$\label{enc:guessRoute0}$ solutionRoute(V, R) :- possibleRouteOfVehicle(V, R), vehicle(V, 0).
$\label{enc:solutionStreet}$ solutionStreet(V, S) :- solutionRoute(V,R), streetOnRoute(S, R,_,_).
$\label{enc:guessEnter}$ 1 {enter(V,S,T) : time(T), T >= MIN, T <= MAX} 1 :- vehicle(V, 1), solutionStreet(V, S), solutionRoute(V, R), streetOnRoute(S, R, MIN, MAX), not origin(V,S).
$\label{enc:enterOrigin}$ enter(V,S,0) :- origin(V,S).
$\label{enc:guessExit}$ 1 {exit(V,S,T) : time(T), T > IN, T <= IN + MAX} 1 :- vehicle(V, 1), enter(V,S,IN), maxTrafficTravelTime(S,MAX).
$\label{enc:nVehicleOnStreet}$ nVehicleOnStreet(S,T,N) :- enter(_,S,T), N = #sum{1,V: enter(V,S,IN), IN <= T; -1,V: exit(V,S,OUT), OUT <= T}.
$\label{enc:travelTimeHeavy}$ travelTime(S,T,X) :- enter(_,S,T), nVehicleOnStreet(S,T,N), heavyTrafficThreshold(S,A,_), N >= A, heavyTrafficTravelTime(S,X).
$\label{enc:travelTimeMedium}$ travelTime(S,T,X) :- enter(_,S,T), nVehicleOnStreet(S,T,N), mediumTrafficThreshold(S,A,B), N >= A, N < B, mediumTrafficTravelTime(S,X).
$\label{enc:travelTimeLight}$ travelTime(S,T,X) :- enter(_,S,T), nVehicleOnStreet(S,T,N), lightTrafficThreshold(S,_,B), N < B, lightTrafficTravelTime(S,X).
$\label{enc:constrTravelTime}$ :- vehicle(V,1), exit(V,S,OUT), enter(V,S,IN), travelTime(S,IN,X), OUT < IN + X.
$\label{enc:constrLink}$ :- vehicle(V,1), exit(V,S1,OUT1), enter(V,S2,IN2), link(S1,S2), IN2 != OUT1.
$\label{enc:constrCapacity}$ :- enter(V,S,T), vehicle(V,1), capacity(S,MAX), nVehicleOnStreet(S,T,N), N > MAX.
$\label{enc:constrRoundabout}$ :- enter(V,SR,T), streetInRoundabout(SR,R), vehicle(V,_), roundabout(R,MAX), #sum{X,S: nVehicleOnStreet(S,T,X), streetInRoundabout(S,R)} = N, N > MAX.
$\label{enc:optCost}$ :~ solutionRoute(V, R), vehicle(V,1), cost(V,R,N). [N@2, V, R]
$\label{enc:optVehicles}$ :~ nVehicleOnStreet(S,T,N). [N@2,S,T]
    \end{asp}
    \caption{ASP Encoding used in the optimized scheduler}
    \label{enc}
\end{figure}

\paragraph{Rules.} Figure \ref{enc} lists the ASP encoding used in the optimized scheduler. Rule $r_{\ref{enc:guessRoute1}}$ defines the atom \ttt{solutionRoute}, which, for every \textit{controlled} vehicle, chooses a single route between all the most different shortest routes which move the vehicle from its origin to its destination, as computed by the preprocessor. Rule $r_{\ref{enc:guessRoute0}}$, instead, imposes the \ttt{solutionRoute} for \textit{simulated} vehicles equal to the route they are already running. Rule $r_{\ref{enc:solutionStreet}}$ defines the atom \ttt{solutionStreet(V,S)} which signals that the vehicle \ttt{V} will run through street \ttt{S} in the solution. Rule $r_{\ref{enc:guessEnter}}$ is used to compute an entry time for every \textit{controlled} vehicle in the minimum and maximum range computed in the preprocessor; for the first origin street, the entry time is imposed to zero with Rule $r_{\ref{enc:enterOrigin}}$. Similarly, rule $r_{\ref{enc:guessExit}}$ defines the exit time of vehicles for every street in their route. Rule $r_{\ref{enc:nVehicleOnStreet}}$ defines the atom \ttt{nVehicleOnStreet(S,T,N)} which counts the number of vehicles on street \ttt{S} at time \ttt{T}. This atom is then used in rules $r_{\ref{enc:travelTimeHeavy}}$ trough $r_{\ref{enc:travelTimeLight}}$  to compute the atom \ttt{travelTime(S,T,X)} which represents the time (\ttt{X}) a vehicle which enters the street \ttt{S} at time \ttt{T} would take to run the whole length of the street in situations of, respectively, heavy, medium or light traffic. Constraint $r_{\ref{enc:constrTravelTime}}$ imposes that when a vehicle enters a street, it cannot leave it before the amount of time specified in the atom \ttt{travelTime}; this defines only a lower-bound, since the vehicle could remain congested and leave much after. Constraint $r_{\ref{enc:constrLink}}$ imposes an order between streets in a route. Constraints $r_{\ref{enc:constrCapacity}}$ and $r_{\ref{enc:constrRoundabout}}$ force the vehicle to respect the capacities of streets and roundabouts, respectively. The weak constraint $r_{\ref{enc:optCost}}$ optimizes the cost of each route for \textit{controlled} vehicles. If two solution produces the same cost, weak constraint $r_{\ref{enc:optVehicles}}$ breaks the tie by preferring the scheduling  which allows vehicles to better spread along the network.

\section{Preliminary Results} \label{sec:results}
\begin{figure}
    \centering
    
\end{figure}

\begin{figure}[htb]
    \centering
    \includegraphics[width=1\textwidth]{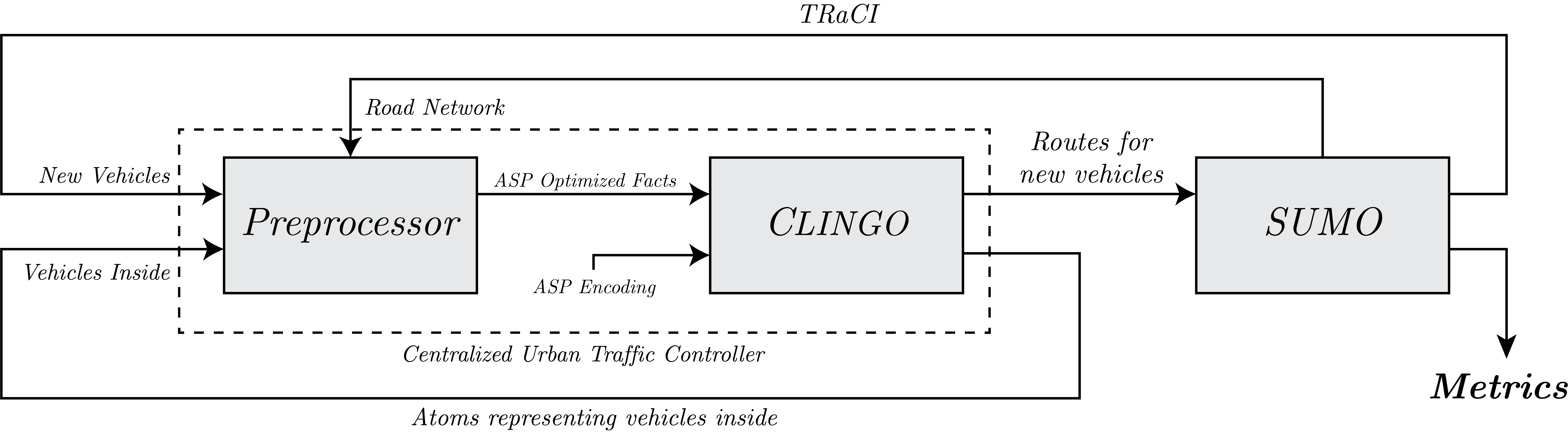}
    \caption{The architecture of the experiments.}
    \label{fig:architecture-specific}
\end{figure}

\begin{figure}[htb]
    \centering
    \includegraphics[width=0.75\textwidth]{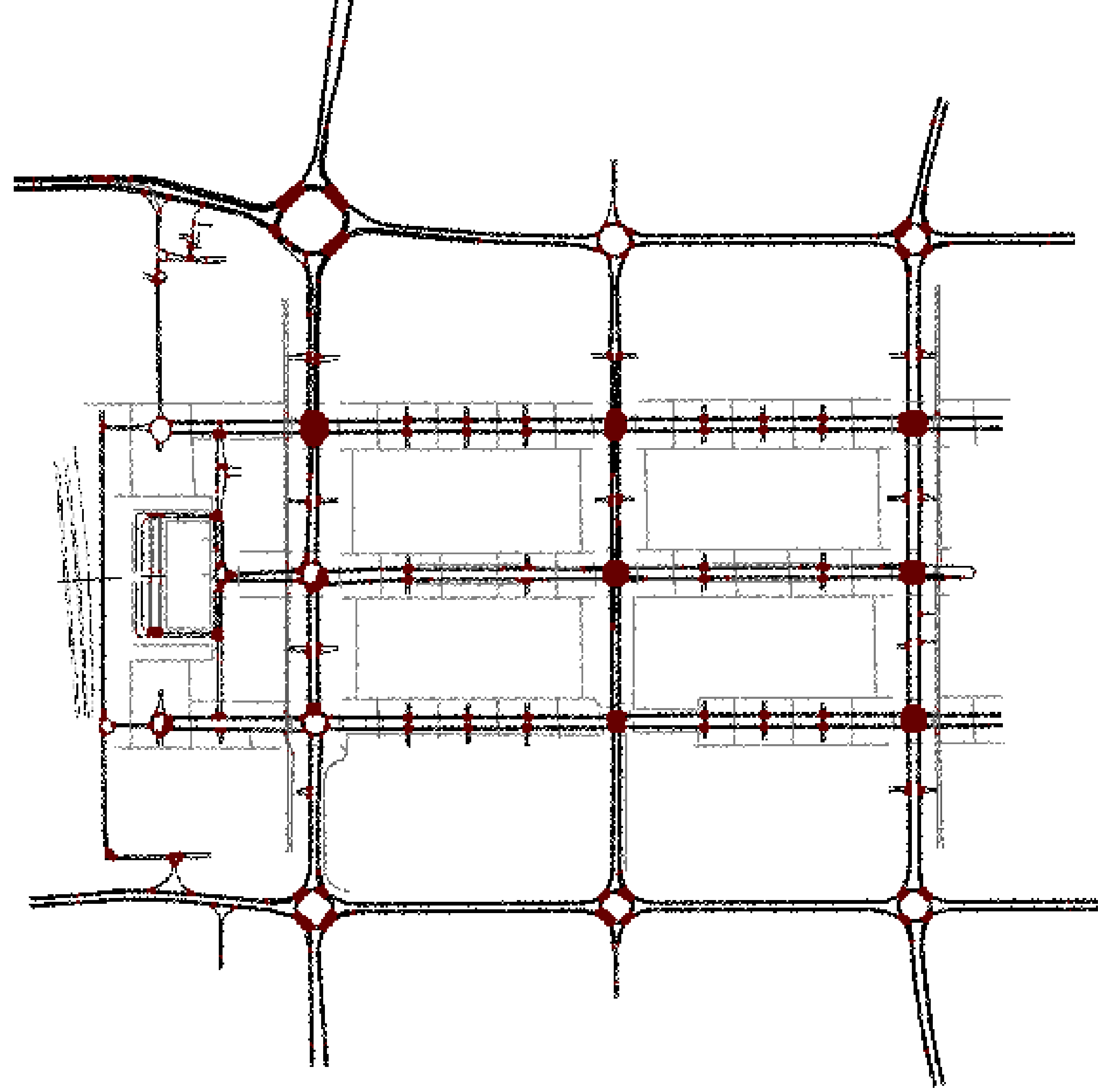}
    \caption{A map of the centre of Milton Keynes (UK), the network in which the experiments are performed}
    \label{fig:milton-keynes}
\end{figure}

\paragraph{Architecture of the experiments.} In order to test the correctness and performance of the proposed approach, it is of paramount importance to connect the system to a real time traffic simulator which can capture all the nuances of traffic flow which are not covered in the modelling (i.e., traffic lights, rights of way, overtakes, etc). Figure \ref{fig:architecture-specific} shows how the CUTC was wired to the traffic simulator SUMO. The traffic control interface TRaCi \cite{wegener2008traci} connects SUMO with the Preprocessor to fetch a representation of the road network and to get notified when new vehicles approach the network and optimized routes needs to be computed; the \textsc{Clingo} wrapper, after computing the optimal routes of approaching vehicles, returns to the preprocessor a list of atoms representing the expected positions of vehicles inside the network, so that, when computing the routes of new approaching vehicles the position of vehicles inside can be accounted for.

The network of Milton Keynes, which is used in our experiments, is shown in Figure \ref{fig:milton-keynes}. Milton Keynes is a town in the United Kingdom, located about $80$ kilometres north-west of London, with a population of approximately $230,000$. The model covers an area of approximately $2.9$ square kilometres, and includes more than $25$ junctions and more than $50$ links.

The experiment presented in this section follow an iterative approach. At time $t=0$, the network is empty and new vehicles approach the network, for these vehicles the CUTC will compute both the route and the expected positions of vehicles in the street (in terms of \ttt{enter} and \ttt{exit} atoms). At the following time step, those vehicles will become \textit{simulated} and the CUTC will then find routes of new approaching vehicles, keeping in consideration the position of \textit{simulated} vehicles found in the previous iterations.

\paragraph{Comparison with real traffic.} For the purpose of evaluating the performance of the proposed framework, in this paragraph, we will compare real traffic data of the Milton Keynes urban area with a simulation in which the same vehicles are routed using our proposed approach. In order to be able to compare the real data with our framework, historical traffic data provided by the Milton Keynes Council and gathered by sensors distributed in the region between December 2015 and December 2016, has been transformed into a SUMO simulation model. The model simulates the morning rush hour (between 8am and 9am on non-holiday weekdays), in which $1900$ vehicles move inside the network. The model has been calibrated and validated. In order to obtain the performance for the ASP-based framework, the architecture shown in Figure \ref{fig:architecture-specific} has been put in place. Table \ref{tab:comparison} shows a comparison between the two approaches in terms of \textit{Total Duration} (i.e. the time the last vehicles exits the network), \textit{Average Route Length, Speed, Duration, Waiting Time} (i.e. the time vehicles spend in queues) and \textit{Depart Delay} (i.e. time vehicles spent waiting for the road to free in order to enter the network). As it can be seen by the comparison, the proposed approach is able to greatly increase the over-all performance of the network, spreading traffic and reducing congestion, increasing the average speed of vehicles and allowing the network to free faster.

\begin{table}[t]
\centering
\begin{tabular}{|l|c|c|}
\hline
 & \textit{Real} & \textit{ASP} \\ \hline
Total Duration {[}s{]} & 15,729 & 5,065 \\ \hline
Avg. Route Length {[}m{]} & 2,465 & 2,107 \\ \hline
Avg. Speed {[}m/s{]} & 2.49 & 5.28 \\ \hline
Avg. Duration {[}s{]} & 3,718.95 & 515.82 \\ \hline
Avg. Waiting Time {[}s{]} & 3,132.36 & 259.39 \\ \hline
% Avg, Time Loss {[}s{]} & 3.514,7 & 362,49 \\ \hline
Avg. Depart Delay [s] & 791.78 & 55.69 \\ \hline
\end{tabular}
\caption{Performance of real traffic data coming from the Milton Keynes urban area and the same vehicles routed using the proposed approach}
\label{tab:comparison}
\end{table}

\paragraph{Scalability.} In the proposed framework, the ASP encoding has to keep track of the position of hundreds of vehicles inside the network in order to route the traffic of new approaching vehicles in an optimal way. In the experiment presented in the previous paragraph, for example, ten minutes after the start of the simulation, up to $200$ vehicles are inside the network. Even with this large number of vehicles, the \textsc{Clingo} solver is able to find an \textit{optimal} route in less than $5s$ in most of the cases. This is possible due to the preliminary work performed by the preprocessor which leaves to the optimizer to explore a pruned search space containing only viable solutions, of which the best must be found. To further improve performance, \textsc{Clingo} is executed with the option \ttt{--parallel-mode=2} in order to parallelize two optimization algorithms: Branch and Bound with Restart on Model \cite{gebser2015progress} and the Unsatisfiable Shrinking Core \cite{alviano2020unsatisfiable} which has already proven to be effective in \cite{DBLP:conf/ruleml/CardelliniNDGGM21}. Experiments were run on a MacBook Pro with a 2.5 GHz Intel Core i7 quad-core, with 16 GB of RAM. The fast resolution of the problems gives the proposed framework the possibility to be implemented in real time scenarios.

\section{Conclusions and Future Work} \label{sec:future}
In this paper, we presented an ASP-based framework which allows to efficiently simulate the flow of traffic in large real-life networks. We showed how this framework can be used to optimize the cost of routes and the time it takes for a vehicle to complete them. A comparison with real data showed that the proposed framework is able to reduce congestion and guide the vehicles in reaching their destination faster. We have also shown how the proposed framework is able to scale in order to be able to deal with instances with hundreds of vehicles inside. Even if the results proved the feasibility of our approach and showed good performances, the proposed framework can still be improved with the aim of increasing its modelling capacities and consequentially the optimization of the traffic flow. 

\paragraph{Future work.} In the following years of his PhD, the author plans to tackle several, more complicated, aspects of Urban Traffic, which would help to improve the modelling capacity of the framework and reduce congestion in the network even further:
\begin{itemize}
    \item Implement other metrics which do not depend directly on the route chosen by the vehicles. In the proposed approach, indeed, only costs associated with the route of the vehicles were considered; this can be used to model a monetary cost of the street (for example tolls) and also to penalize certain routes for certain types of vehicles (we could suppose that trucks or heavy vehicles should avoid entering the city centre). On the other hand, pollution is more correlated with the number of vehicles, with their speed and with the number of \textit{"start and stops"} the vehicles take.
    \item Deal with re-routing of vehicles inside the network. In our proposed approach, vehicle routing is only possible when the vehicle enters the network. While this approach is easier to implement and requires fewer communications between vehicles and the CUTC can lead to suboptimal solutions. In fact, in some scenarios, it would also be beneficial to reroute vehicles inside the network with the aim to adjust traffic and allow a more rapid flow of vehicles.
    \item Model traffic lights and keep track of their switch phases in order to route traffic accordingly. The presence of traffic lights could be modelled in two ways: (i) by simply adding a penalizing time (i.e., the average time in which the traffic light is red) for routes which includes traffic lights, or (ii) by modelling precisely the time dependent switching of green phases of traffic lights in the ASP encoding. These two methods differ enormously in terms of complexity, but are both worth exploring in order to investigate their benefits.
    \item Model larger and more complicated urban areas. We discussed how the proposed approach is able to rapidly find optimal solutions in scenarios with up to $200$ vehicles. Even if the proposed approach can successfully manage traffic in a small network like Milton Keynes, it would be interesting to investigate the solution on larger urban areas with thousands of vehicles.
\end{itemize}
\bibliographystyle{eptcs}
\bibliography{main}
\end{document}